\documentclass[11pt]{article}
\usepackage{coling2018}
\usepackage{times}
\usepackage{url}
\usepackage{latexsym}
\usepackage{graphicx}
\usepackage{multirow}
\usepackage{amsmath}
\usepackage{amsfonts}
\usepackage{color}
\usepackage{tabularx}

\makeatletter 
  \newcommand\figcaption{\def\@captype{figure}\caption} 
  \newcommand\tabcaption{\def\@captype{table}\caption} 
\makeatother

\newcommand{\para}[1]{\vspace{0.05in}\noindent\textbf{#1 }}

\newcolumntype{C}[1]{>{\centering\arraybackslash}p{#1}}

\title{Neural Collective Entity Linking}

\author{Yixin Cao, Lei Hou\thanks{Corresponding author.}, Juanzi Li, Zhiyuan Liu\\
Dept. of Computer Science and Technology, Tsinghua University, China 100084 \\
{\tt \{caoyixin2011,greener2009,lijuanzi2008\}@gmail.com}\\
{\tt liuzy@tsinghua.edu.cn} \\}

\date{}

\begin{document}
\maketitle
\begin{abstract}
Entity Linking aims to link entity mentions in texts to knowledge bases, and neural models have achieved recent success in this task. However, most existing methods rely on local contexts to resolve entities independently, which may usually fail due to the data sparsity of local information. To address this issue, we propose a novel neural model for collective entity linking, named as NCEL. NCEL applies Graph Convolutional Network to integrate both local contextual features and global coherence information for entity linking. To improve the computation efficiency, we approximately perform graph convolution on a subgraph of adjacent entity mentions instead of those in the entire text. We further introduce an attention scheme to improve the robustness of NCEL to data noise and train the model on Wikipedia hyperlinks to avoid overfitting and domain bias. In experiments, we evaluate NCEL on five publicly available datasets to verify the linking performance as well as generalization ability. We also conduct an extensive analysis of time complexity, the impact of key modules, and qualitative results, which demonstrate the effectiveness and efficiency of our proposed method.
\end{abstract}

\section{Introduction}
\label{sec:intro}

\blfootnote{\hspace{-0.65cm}  
    This work is licensed under a Creative Commons Attribution 4.0 International License. License details: \url{http://creativecommons.org/licenses/by/4.0/}
}

Entity linking (EL), mapping entity mentions in texts to a given knowledge base (KB), serves as a fundamental role in many fields, such as question answering~\cite{Zhang2016AJM}, semantic search~\cite{Blanco2015FastAS}, and information extraction~\cite{Ji2015OverviewOT,ji2016overview}. However, this task is non-trivial because entity mentions are usually ambiguous. As shown in Figure~\ref{fig:example}, the mention \textit{England} refers to three entities in KB, and an entity linking system should be capable of identifying the correct entity as \textit{England cricket team} rather than \textit{England} and \textit{England national football team}.

Entity linking is typically broken down into two main phases: (i) candidate generation obtains a set of referent entities in KB for each mention, and (ii) named entity disambiguation selects the possible candidate entity by solving a ranking problem. The key challenge lies in the ranking model that computes the relevance between candidates and the corresponding mentions based on the information both in texts and KBs~\cite{Nguyen2016JointLO}. In terms of the features used for ranking, we classify existing EL models into two groups: \textbf{local models} to resolve mentions independently relying on textual context information from the surrounding words~\cite{Chen2011CollaborativeRA,Chisholm2015EntityDW,Lazic2015PlatoAS,Yamada2016JointLO}, and \textbf{global (collective) models}, which are the main focus of this paper, that encourage the target entities of all mentions in a document to be topically coherent~\cite{Han2011CollectiveEL,Cassidy2012AnalysisAE,He2013EfficientCE,Cheng2013RelationalIF,Durrett2014AJM,Huang2014CollectiveTW}.

Global models usually build an entity graph based on KBs to capture coherent entities for all identified mentions in a document, where the nodes are entities, and edges denote their relations. The graph provides highly discriminative semantic signals (e.g., entity relatedness) that are unavailable to local model~\cite{Eshel2017NamedED}. For example (Figure~\ref{fig:example}), an EL model seemly cannot find sufficient disambiguation clues for the mention \textit{England} from its surrounding words, unless it utilizes the coherence information of consistent topic ``cricket" among adjacent mentions \textit{England}, \textit{Hussain}, and \textit{Essex}. Although the global model has achieved significant improvements, its limitation is threefold:

\begin{figure}[htb]
\centerline{\includegraphics[width=\textwidth]{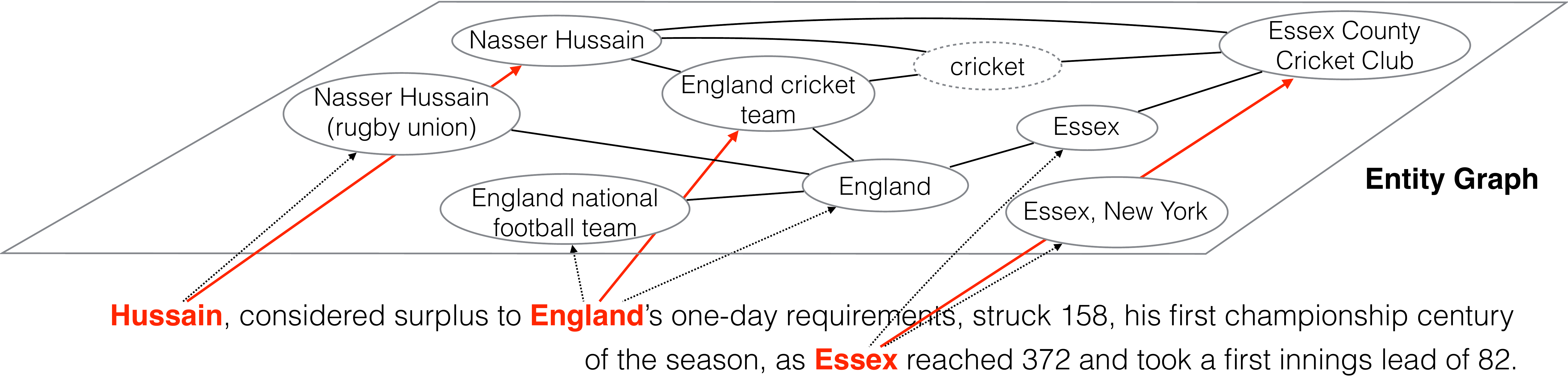}}
\caption{Illustration of named entity disambiguation for three mentions \textit{England}, \textit{Hussain}, and \textit{Essex}. The nodes linked by arrowed lines are the candidate entities, where red solid lines denote target entities.}
\label{fig:example}
\end{figure}

\begin{enumerate}
\item The global approach suffers from the data sparsity issue of unseen words/entities, and the failure to induce underlying discriminative features for EL.
\item The joint inference mechanism in the global approach leads to expensive computations, especially when the entity graph may contain hundreds of nodes in case of long documents.
\item The annotated EL training data is usually expensive to obtain or only available in narrow domains, which results in possible overfitting issue or domain bias.
\end{enumerate}

To mitigate the first limitation, recent EL studies introduce neural network (NN) models due to its amazing feature abstraction and generalization ability. In such models, words/entities are represented by low dimensional vectors in a continuous space, and features for mention as well as candidate entities are automatically learned from data~\cite{Nguyen2016JointLO}. However, existing NN-based methods for EL are either local models~\cite{Yamada2017LearningDR,Gupta2017EntityLV} or merely use word/entity embeddings for feature extraction and rely on another modules for collective disambiguation, which thus cannot fully utilize the power of NN models for collective EL~\cite{Globerson2016CollectiveER,Barbosa2017RobustNE,Phan2018PairLinkingFC}.

The second drawback of the global approach has been alleviated through approximate optimization techniques, such as PageRank/random walks~\cite{Pershina2015PersonalizedPR}, graph pruning~\cite{hoffart2011robust}, ranking SVMs~\cite{Ratinov2011LocalAG}, or loopy belief propagation (LBP)~\cite{Globerson2016CollectiveER,Ganea2017DeepJE}. However, these methods are not differentiable and thus difficult to be integrated into neural network models (the solution for the first limitation).

To overcome the third issue of inadequate training data, \cite{Gupta2017EntityLV} has explored a massive amount of hyperlinks in Wikipedia, but these potential annotations for EL contain much noise, which may distract a naive disambiguation model~\cite{Chisholm2015EntityDW}.

In this paper, we propose a novel \textbf{N}eural \textbf{C}ollective \textbf{E}ntity \textbf{L}inking model (NCEL), which performs global EL combining deep neural networks with Graph Convolutional Network (GCN)~\cite{Defferrard2016ConvolutionalNN,Kipf2016SemiSupervisedCW} that allows flexible encoding of entity graphs. It integrates both local contextual information and global interdependence of mentions in a document, and is efficiently trainable in an end-to-end fashion. Particularly, we introduce attention mechanism to robustly model local contextual information by selecting informative words and filtering out the noise. On the other hand, we apply GCNs to improve discriminative signals of candidate entities by exploiting the rich structure underlying the correct entities. To alleviate the global computations, we propose to convolute on the subgraph of adjacent mentions. Thus, the overall coherence shall be achieved in a chain-like way via a sliding window over the document. To the best of our knowledge, this is the first effort to develop a unified model for neural collective entity linking.

In experiments, we first verify the efficiency of NCEL via theoretically comparing its time complexity with other collective alternatives. Afterwards, we train our neural model using collected Wikipedia hyperlinks instead of dataset-specific annotations, and perform evaluations on five public available benchmarks. The results show that NCEL consistently outperforms various baselines with a favorable generalization ability. Finally, we further present the performance on a challenging dataset WW~\cite{Barbosa2017RobustNE} as well as qualitative results, investigating the effectiveness of each key module.

\section{Preliminaries and Framework}
We denote $M=\{m_i\}$ as a set of entity mentions in a document $D=\langle x_1,\dots,x_i,\dots,x_{|D|}\rangle$, where $x_i$ is either a word $w_i$ or a mention $m_i$. $G=(E,R)$ is the entity graph for document $D$ derived from the given knowledge base, where $E=\{e_i\}$ is a set of entities, $R=\{r^i_j\in(0,1]\}$ denotes the relatedness between $\langle e_i,e_j\rangle$ and higher values indicate stronger relations. Based on $G$, we extract a subgraph $G^{ij}$ for $e_j\in\Phi(m_i)$, where $\Phi(m_i)$ denotes the set of candidate entities for $m_i$. Note that we don't include the relations among candidates of the same mention in $G^{ij}$ because these candidates are mutually exclusive in disambiguation.

Formally, we define the entity linking problem as follows: Given a set of mentions $M$ in a document $D$, and an entity graph $G$, the goal is to find an assignment\footnote{Normally, an entity linking system outputs NIL for a mention when no assignment score is higher than a threshold. This is application-specific and thus outside of the scope of this work.} $\Gamma:M\rightarrow E$.

\begin{figure}[h]
\centerline{\includegraphics[width=\textwidth]{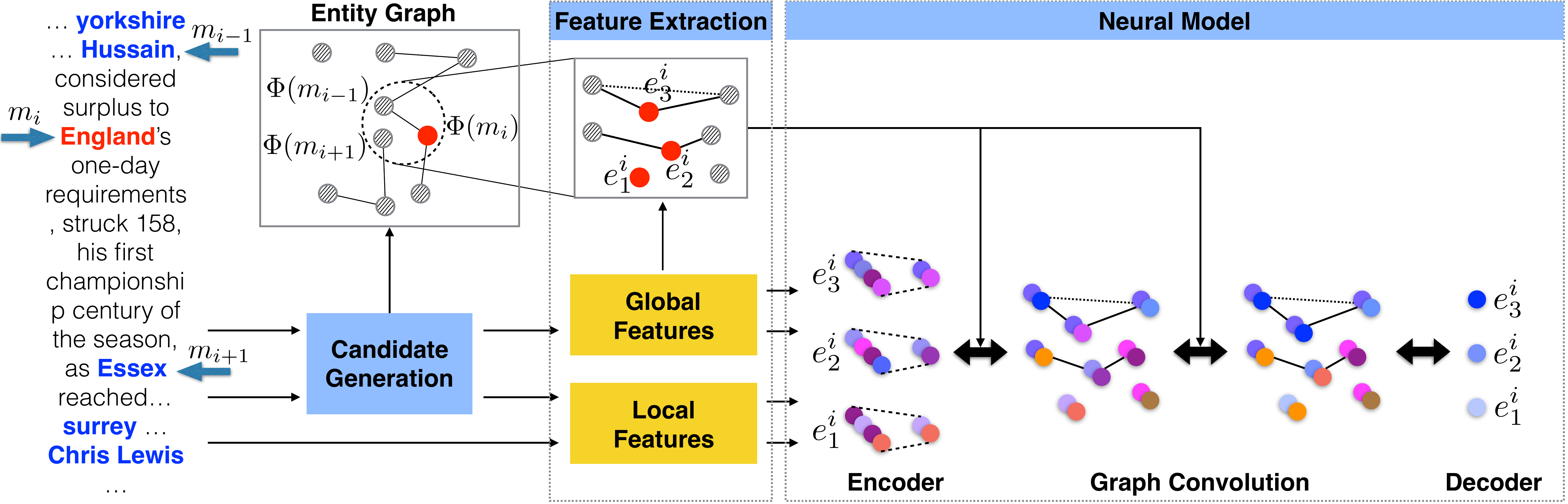}}
\caption{Framework of NCEL. The inputs of a set of mentions in a document are listed in the left side. The words in red indicate the current mention $m_i$, where $m_{i-1},m_{i+1}$ are neighbor mentions, and $\Phi(m_i)=\{e_1^i,e_2^i,e_3^i\}$ denotes the candidate entity set for $m_i$.}
\label{fig:frame}
\end{figure}

To collectively find the best assignment, NCEL aims to improve the discriminability of candidates' local features by using entity relatedness within a document via GCN, which is capable of learning a function of features on the graph through shared parameters over all nodes. Figure~\ref{fig:frame} shows the framework of NCEL including three main components:

\begin{enumerate}
\item \textbf{Candidate Generation}: we use a pre-built dictionary to generate a set of entities as candidates to be disambiguated for each mention, e.g., for mention \textit{England}, we have $\Phi(m_i)=\{e_1^i,e_2^i,e_3^i\}$, in which the entities refer to \textit{England national football team}, \textit{England} and \textit{England cricket team} in Figure~\ref{fig:example}, respectively.
\item \textbf{Feature Extraction}: based on the document and its entity graph, we extract both local features and global features for each candidate entity to feed our neural model. Concretely, local features reflect the compatibility between a candidate and its mention within the contexts, and global features are to capture the topical coherence among various mentions. These features, including vectorial representations of candidates and a subgraph indicating their relatedness, are highly discriminative for tackling ambiguity in EL.
\item \textbf{Neural Model}: given feature vectors and subgraphs of candidates, we first encode the features to represent nodes (i.e., candidates) in the graph, then improve them for disambiguation via multiple graph convolutions by exploiting the structure information, in which the features for correct candidates that are strongly connected (i.e., topical coherent) shall enhance each other, and features for incorrect candidates are weakened due to their sparse relations. Finally, we decode the features of nodes to output a probability indicating how possible the candidate refers to its mention.
\end{enumerate}

\noindent\textbf{Example} As shown in Figure~\ref{fig:frame}, for the current mention \textit{England}, we utilize its surrounding words as local contexts (e.g., \textit{surplus}), and adjacent mentions (e.g., \textit{Hussian}) as global information. Collectively, we utilize the candidates of \textit{England} $e_j^i\in\Phi(m_i),j=1,2,3$ as well as those entities of its adjacencies $\Phi(m_{i-1})\cup\Phi(m_{i+1})$ to construct feature vectors for $e_j^i$ and the subgraph of relatedness as inputs of our neural model. Let darker blue indicate higher probability of being predicted, the correct candidate $e_3^i$ becomes bluer due to its bluer neighbor nodes of other mentions $m_{i-1},m_{i+1}$. The dashed lines denote entity relations that have indirect impacts through the sliding adjacent window , and the overall structure shall be achieved via multiple sub-graphs by traversing all mentions. 

Before introducing our model, we first describe the component of candidate generation.

\subsection{Candidate Generation}
\label{sec:gen}
Similar to previous work~\cite{Ganea2017DeepJE}, we use the prior probability $\hat{p}(e_i|m_j)$ of entity $e_i$ conditioned on mention $m_j$ both as a local feature and to generate candidate entities: $\Phi(m_j)=\{e_i|\hat{p}(e_i|m_j)>0\}$. We compute $\hat{p}(\cdot)$ based on statistics of mention-entity pairs from: (i) Wikipedia page titles, redirect titles and hyperlinks, (ii) the dictionary derived from a large Web Corpus~\cite{Spitkovsky2012ACD}, and (iii) the YAGO dictionary with a uniform distribution~\cite{hoffart2011robust}. We pick up the maximal prior if a mention-entity pair occurs in different resources. In experiments, to optimize for memory and run time, we keep only top $n$ entities based on $\hat{p}(e_i|m_j)$. In the following two sections, we will present the key components of NECL, namely feature extraction and neural network for collective entity linking.

\section{Feature Extraction}
The main goal of NCEL is to find a solution for collective entity linking using an end-to-end neural model, rather than to improve the measurements of local textual similarity or global mention/entity relatedness. Therefore, we use joint embeddings of words and entities at sense level~\cite{Cao2017BridgeTA} to represent mentions and its contexts for feature extraction. In this section, we give a brief description of our embeddings followed by our features used in the neural model.

\subsection{Learning Joint Embeddings of Word and Entity}
\label{sec:emb}
Following~\cite{Cao2017BridgeTA}, we use Wikipedia articles, hyperlinks, and entity outlinks to jointly learn word/mention and entity embeddings in a unified vector space, so that similar words/mentions and entities have similar vectors. To address the ambiguity of words/mentions, \cite{Cao2017BridgeTA} represents each word/mention with multiple vectors, and each vector denotes a sense referring to an entity in KB. The quality of the embeddings is verified on both textual similarity and entity relatedness tasks.

Formally, each word/mention has a global embedding $\mathbf{w}_i/\mathbf{m}_i$, and multiple sense embeddings $\mathcal{S}(x_i)=\{\mathbf{s}_j\}$. Each sense embedding $\mathbf{s}_j$ refers to an entity embedding $\mathbf{e}_j$, while the difference between $\mathbf{s}_j$ and $\mathbf{e}_j$ is that $\mathbf{s}_j$ models the co-occurrence information of an entity in texts (via hyperlinks) and $\mathbf{e}_j$ encodes the structured entity relations in KBs. More details can be found in the original paper.

\subsection{Local Features}
\label{sec:lf}
Local features focus on how compatible the entity is mentioned in a piece of text (i.e., the mention and the context words). Except for the prior probability (Section~\ref{sec:gen}), we define two types of local features for each candidate entity $e_j\in\Phi(m_i)$:

\para{String Similarity} Similar to~\cite{Yamada2017LearningDR}, we define string based features as follows: the edit distance between mention's surface form and entity title, and boolean features indicating whether they are equivalent, whether the mention is inside, starts with or ends with entity title and vice versa.

\para{Compatibility} We also measure the compatibility of $e_j$ with the mention's context words $\mathcal{C}(m_i)$ by computing their similarities based on joint embeddings: $sim(\mathbf{e}_j,\mathbf{c}_{m_i,e_j})$ and $sim(\mathbf{s}_j,\mathbf{c}_{m_i,e_j})$, where $\mathbf{c}_{m_i,e_j}$ is the context embedding of $m_i$ conditioned on candidate $e_j$ and is defined as the average sum of word global vectors weighted by attentions:

$$\mathbf{c}_{m_i,e_j}=\sum_{w_k\in \mathcal{C}(m_i)}\alpha_{kj}\mathbf{w}_k$$
where $\alpha_{kj}$ is the $k$-th word's attention from $e_j$. In this way, we automatically select informative words by assigning higher attention weights, and filter out irrelevant noise through small weights. The attention $\alpha_{kj}$ is computed as follows:

$$\alpha_{kj}\propto sim(\mathbf{w}_k,\mathbf{e}_j)$$
where $sim$ is the similarity measurement, and we use cosine similarity in the presented work. We concatenate the prior probability, string based similarities, compatibility similarities and the embeddings of contexts as well as the entity as the local feature vectors.

\subsection{Global Features}
\label{sec:gf}
The key idea of collective EL is to utilize the topical coherence throughout the entire document. The consistency assumption behind it is that: \textit{all mentions in a document shall be on the same topic}. However, this leads to exhaustive computations if the number of mentions is large. Based on the observation that the consistency attenuates along with the distance between two mentions, we argue that the adjacent mentions might be sufficient for supporting the assumption efficiently.

Formally, we define neighbor mentions as $q$ adjacent mentions before and after current mention $m_i$: $\mathcal{N}(m_i)=\{m_{i-q},\dots,m_{i-1},m_{i+1},\dots,m_{i+q}\}$, where $2q$ is the pre-defined window size. Thus, the topical coherence at document level shall be achieved in a chain-like way. As shown in Figure~\ref{fig:frame} ($q=1$), mentions \textit{Hussain} and \textit{Essex}, a cricket player and the cricket club, provide adequate disambiguation clues to induce the underlying topic ``cricket" for the current mention \textit{England}, which impacts positively on identifying the mention \textit{surrey} as another cricket club via the common neighbor mention \textit{Essex}.

A degraded case happens if $q$ is large enough to cover the entire document, and the mentions used for global features become the same as the previous work, such as~\cite{Pershina2015PersonalizedPR}. In experiments, we heuristically found a suitable $q=3$ which is much smaller than the total number of mentions. The benefits of efficiency are in two ways: (i) to decrease time complexity, and (ii) to trim the entity graph into a fixed size of subgraph that facilitates computation acceleration through GPUs and batch techniques, which will be discussed in Section~\ref{sec:complexity}.

Given neighbor mentions $\mathcal{N}(m_i)$, we extract two types of vectorial global features and structured global features for each candidate $e_j\in\Phi(m_i)$:

\para{Neighbor Mention Compatibility} Suppose neighbor mentions are topical coherent, a candidate entity shall also be compatible with neighbor mentions if it has a high compatibility score with the current mention, otherwise not. That is, we extract the vectorial global features by computing the similarities between $e_j$ and all neighbor mentions: $\{sim(\mathbf{e}_j,\mathbf{m}_i)|m_i\in\mathcal{N}(m_i)\}$, where $\mathbf{m}_i$ is the mention embedding by averaging the global vectors of words in its surface form: 
$\mathbf{m}_j=\sum_{w_l\in\mathcal{T}(m_j)}\mathbf{w}_l$, 
where $\mathcal{T}(m_j)$ are tokenized words of mention $m_j$.

\para{Subgraph Structure} The above features reflect the consistent semantics in texts (i.e., mentions). We now extract structured global features using the relations in KB, which facilitates the inference among candidates to find the most topical coherent subset. For each document, we obtain the entity graph $G$ by taking candidate entities of all mentions $\Phi(M)$ as nodes, and using entity embeddings to compute their similarities as edges $R=\{r^i_j|r^i_j=sim(\mathbf{e}_i,\mathbf{e}_j)\}$. Then, we extract the subgraph structured features $\mathbf{g}^{i*}$ for each entity $e_*^i\in\Phi(m_i),m_i\in M$ for efficiency.

Formally, we define the subgraph as: $G^{i*}=(e_*^i\bigcup\Phi(\mathcal{N}(m_i)), R^{i*})$, where $R^{i*}=\{r^{i*}_{jk}|e^j_k\in\Phi(m_j),j\in[i-q,i+q]\setminus i\}$. For example (Figure~\ref{fig:example}), for entity \textit{England cricket team}, the subgraph contains the relation from it to all candidates of neighbor mentions: \textit{England cricket team}, \textit{Nasser Hussain (rugby union)}, \textit{Nasser Hussain}, \textit{Essex}, \textit{Essex County Cricket Club} and \textit{Essex, New York}. To support batch-wise acceleration, we represent $G^{i*}$ in the form of adjacency table based vectors: $\mathbf{g}^{i*}=[r^{i*}_{i-q,1},\cdots,r^{i*}_{i+q,n}]^T\in \mathbb{R}^{2qn}$, where $n$ is the number of candidates per mention.

Finally, for each candidate $e^i_j$, we concatenate local features and neighbor mention compatibility scores as the feature vector $\mathbf{f}^{ij}$, and construct the subgraph structure representation $\mathbf{g}^{ij}$ as the inputs of NCEL.

\section{Neural Collective Entity Linking}
\label{sec:ncel}
NCEL incorporates GCN into a deep neural network to utilize structured graph information for collectively feature abstraction, while differs from conventional GCN in the way of applying the graph. Instead of the entire graph, only a subset of nodes is ``visible" to each node in our proposed method, and then the overall structured information shall be reached in a chain-like way. Fixing the size of the subset, NCEL is further speeded up by batch techniques and GPUs, and is efficient to large-scale data.

\subsection{Graph Convolutional Network}
GCNs are a type of neural network model that deals with structured data. It takes a graph as an input and output labels for each node. As a simplification of spectral graph convolutions, the main idea of \cite{Kipf2016SemiSupervisedCW} is similar to a propagation model: to enhance the features of a node according to its neighbor nodes. The formulation is as follows:

$$H^{l+1}=\sigma(\tilde{A}H^lW^l)$$
where $\tilde{A}$ is a normalized adjacent matrix of the input graph with self-connection, $H^l$ and $W^l$ are the hidden states and weights in the $l$-th layer, and $\sigma(\cdot)$ is a non-linear activation, such as \textit{ReLu}.

\subsection{Model Architecture}
As shown in Figure~\ref{fig:frame}, NCEL identifies the correct candidate $e_1^i$ for the mention $m_i$ by using vectorial features as well as structured relatedness with candidates of neighbor mentions $\Phi(m_{i-1}),\Phi(m_{i+1})$. Given feature vector $\mathbf{f}^{ij}\in\mathbb{R}^{d_0}$ and subgraph representation $\mathbf{g}^{ij}\in\mathbb{R}^{2qn}$ of each candidate $e^i_j\in\Phi(m_i)$, we stack them as inputs for mention\footnote{For clarity, we omit the superscripts indicating the mention.} $m_i$: $\mathbf{f}=[\mathbf{f}_{i1},\cdots,\mathbf{f}_{in}]^T\in\mathbb{R}^{n\times d_0}$, and the adjacent matrix $A=[\hat{\mathbf{g}}^1,\cdots,\hat{\mathbf{g}}_n]^T\in\mathbb{R}^{n\times (2qn+1)}$, where $\hat{\mathbf{g}}^j=[\mathbf{g}^j,1]^T\in\mathbb{R}^{2qn+1}$ denotes the subgraph with self-connection. We normalize $A$ such that all rows sum to one, denoted as $\tilde{A}$, avoiding the change in the scale of the feature vectors.

Given $\mathbf{f}$ and $\tilde{A}$, the goal of NCEL is to find the best assignment:

\begin{displaymath}
\Gamma^*(m_i)=\underset{\hat{y}}{\text{argmax}}\ P(\hat{y};\mathbf{f},\tilde{A},\omega)
\end{displaymath}
where $\hat{y}$ is the output variable of candidates, and $P(\cdot)$ is a probability function as follows:

$$P(\hat{y};\mathbf{f},\tilde{A},\omega)\propto \exp(F(\mathbf{f},\tilde{A},\hat{y};\omega))$$
where $F(\mathbf{f},\tilde{A},\hat{y};\omega)$ is the score function parameters by $\omega\in\mathbb{R}^\omega$. NCEL learns the mapping $F(\cdot)$ through a neural network including three main modules: encoder, sub-graph convolution network (sub-GCN) and decoder. Next, we introduce them in turn.

\para{Encoder} The function of this module is to integrate different features by a multi-layer perceptron (MLP):

$$h^1=\sigma(\mathbf{f}W^1+b^1)$$
where $h^1$ is the hidden states of the current mention, $W^1\in\mathbb{R}^{d_0\times d_1}$ and $b^1\in\mathbb{R}^{d_1}$ are trainable parameters and bias. We use ReLu as the non-linear activation $\sigma(\cdot)$.

\para{Sub-Graph Convolution Network} Similar to GCN, this module learns to abstract features from the hidden state of the mention itself as well as its neighbors. Suppose $h^t_{m_k}$ is the hidden states of the neighbor $m_k$, we stack them to expand the current hidden states of $m_i$ as $\tilde{h}^t\in\mathbb{R}^{(2qn+1)\times d_t}$, such that each row corresponds to that in the subgraph adjacent matrix $\tilde{A}$. We define sub-graph convolution as:

$$h^{t+1}=\sigma(\tilde{A}\tilde{h}^tW^t)$$
where $W^t\in\mathbb{R}^{d_t\times d_{t+1}}$ is a trainable parameter.

\para{Decoder} After $T$ iterations of sub-graph convolution, the hidden states integrate both features of $m_i$ and its neighbors. A fully connected decoder maps $h^{t+1}$ to the number of candidates as follows:

$$F=h^{T+1}W^{T+1}$$
where $W^{T+1}\in\mathbb{R}^n$.

\subsection{Training}
The parameters of network are trained to minimize cross-entropy of the predicted and ground truth $y^g$:

$$\mathcal{L}_m=-\sum_{j=1}^n y^g_j log(P(\hat{y}=e_j;\mathbf{f},\tilde{A},\omega))$$

Suppose there are $D\in\mathcal{D}$ documents in training corpus, each document has a set of mentions $M$, leading to totally $M\in\mathcal{M}$ mention sets. The overall objective function is as follows:

$$\mathcal{L}=\sum_{M\in\mathcal{M}}\sum_{m\in M}\mathcal{L}_m$$

\section{Experiments}
\label{sec:exp}
To avoid overfitting with some dataset, we train NCEL using collected Wikipedia hyperlinks instead of specific annotated data. We then evaluate the trained model on five different benchmarks to verify the linking precision as well as the generalization ability. Furthermore, we investigate the effectiveness of key modules in NCEL and give qualitative results for comprehensive analysis\footnote{Our codes can be found in \url{https://github.com/TaoMiner/NCEL}}.

\subsection{Baselines and Datasets}
\label{sec:data}
We compare NCEL with the following state-of-the-art EL methods including three local models and three types of global models:

\begin{enumerate}
\item Local models: He~\cite{He2013LearningER} and Chisholm~\cite{Chisholm2015EntityDW} beat many global models by using auto-encoders and web links, respectively, and NTEE~\cite{Yamada2017LearningDR} achieves the best performance based on joint embeddings of words and entities.

\item Iterative model: AIDA~\cite{hoffart2011robust} links entities by iteratively finding a dense subgraph.

\item Loopy Belief Propagation: Globerson~\cite{Globerson2016CollectiveER} and PBoH~\cite{Ganea2016ProbabilisticBM} introduce LBP~\cite{Murphy1999LoopyBP} techniques for collective inference, and Ganea~\cite{Ganea2017DeepJE} solves the global training problem via truncated fitting LBP.

\item PageRank/Random Walk: Boosting~\cite{Kulkarni2009CollectiveAO}, AGDISTISG~\cite{Usbeck2014AGDISTISG}, Babelfy~\cite{Moro2014EntityLM}, WAT~\cite{Piccinno2014FromTT}, xLisa~\cite{Zhang2014XLiSACS} and WNED~\cite{Barbosa2017RobustNE} performs PageRank~\cite{page1999pagerank} or random walk~\cite{Tong2006FastRW} on the mention-entity graph and use the convergence score for disambiguation.

\end{enumerate}

For fairly comparison, we report the original scores of the baselines in the papers. Following these methods, we evaluate NCEL on the following five datasets: (1) \textbf{CoNLL-YAGO}~\cite{hoffart2011robust}: the CoNLL 2003 shared task including testa of 4791 mentions in 216 documents, and testb of 4485 mentions in 213 documents. (2) \textbf{TAC2010}~\cite{Ji2010OverviewOT}: constructed for the Text Analysis Conference that comprises 676 mentions in 352 documents for testing. (3) \textbf{ACE2004}~\cite{Ratinov2011LocalAG}: a subset of ACE2004 co-reference documents including 248 mentions in 35 documents, which is annotated by Amazon Mechanical Turk. (4) \textbf{AQUAINT}~\cite{Milne2008LearningTL}: 50 news articles including 699 mentions from three different news agencies. (5) \textbf{WW}~\cite{Barbosa2017RobustNE}: a new benchmark with balanced prior distributions of mentions, leading to a hard case of disambiguation. It has 6374 mentions in 310 documents automatically extracted from Wikipedia.


\subsection{Training Details and Running Time Analysis}
\label{sec:complexity}

\para{Training} We collect 50,000 Wikipedia articles according to the number of its hyperlinks as our training data. For efficiency, we trim the articles to the first three paragraphs leading to 1,035,665 mentions in total. Using CoNLL-Test A as the development set, we evaluate the trained NCEL on the above benchmarks. We set context window to 20, neighbor mention window to 6, and top $n=10$ candidates for each mention. We use two layers with 2000 and 1 hidden units in MLP encoder, and 3 layers in sub-GCN. We use early stop and fine tune the embeddings. With a batch size of 16, nearly 3 epochs cost less than 15 minutes on the server with 20 core CPU and the GeForce GTX 1080Ti GPU with 12Gb memory. We use standard Precision, Recall and F1 at mention level (Micro) and at the document level (Macro) as measurements.

\para{Complexity Analysis} Compared with local methods, the main disadvantage of collective methods is high complexity and expensive costs. Suppose there are $k$ mentions in documents on average, among these global models, NCEL not surprisingly has the lowest time complexity $\mathcal{O}(T*kn^2)$ since it only considers adjacent mentions, where $T$ is the number of sub-GCN layers indicating the iterations until convergence. AIDA has the highest time complexity $k^3n^3$ in worst case due to exhaustive iteratively finding and sorting the graph. The LBP and PageRank/random walk based methods achieve similar high time complexity of $\mathcal{O}(T*k^2n^2)$ mainly because of the inference on the entire graph.

\subsection{Results on GERBIL}
\label{sec:gerbil}

GERBIL~\cite{usbeck2015gerbil} is a benchmark entity annotation framework that aims to provide a unified comparison among different EL methods across datasets including ACE2004, AQUAINT and CoNLL. We compare NCEL with the global models that report the performance on GERBIL.

\begin{table}[htp]
\small
\begin{center}
\begin{tabular}{|c|c|c|c|c|c|c|c||c|}
\hline
\textbf{Datasets} & \textbf{AGDISTIS} & \textbf{AIDA} & \textbf{Babelfy} & \textbf{WAT} & \textbf{xLisa} & \textbf{PBoH} & \textbf{WNED} & \textbf{NCEL} \\
\hline
\multirow{2}{*}{\textbf{ACE2004}} & 0.66 & 0.80 & 0.61 & 0.76 & 0.81 & 0.79 & 0.81 & \textbf{0.88} \\
 & 0.78 & 0.89 & 0.76 & 0.85 & 0.88 & 0.86 & \textbf{0.90} & 0.89 \\
\hline
\multirow{2}{*}{\textbf{AQUAINT}} & 0.73 & 0.57 & 0.70 & 0.75 & 0.79 & 0.84 & 0.83 & \textbf{0.87} \\
 & 0.59 & 0.56 & 0.70 & 0.76 & 0.77 & 0.83 & 0.83 & \textbf{0.88} \\
\hline
\multirow{2}{*}{\textbf{CoNLL-Test A}} & 0.56 & 0.74 & 0.74 & 0.78 & 0.52 & \textbf{0.80} & 0.79 & 0.79 \\
 & 0.49 & 0.71 & 0.68 & 0.76 & 0.48 & \textbf{0.77} & 0.76 & \textbf{0.77} \\
\hline
\multirow{2}{*}{\textbf{CoNLL-Test B}} & 0.55 & 0.77 & 0.76 & \textbf{0.80} & 0.54 & \textbf{0.80} & 0.79 & \textbf{0.80} \\
 & 0.54 & 0.78 & 0.70 & \textbf{0.80} & 0.53 & 0.79 & 0.79 & \textbf{0.80} \\
 \hline
\multirow{2}{*}{\textbf{Average}} & 0.63 & 0.72 & 0.70 & 0.77 & 0.67 & 0.81 & 0.81 & \textbf{0.84} \\
 & 0.60 & 0.74 & 0.71 & 0.79 & 0.67 & 0.81 & 0.82 & \textbf{0.84} \\
\hline
\end{tabular}
\caption{Micro F1 (above) and Macro F1 (bottom) on GERBIL.}
\label{tab:gerbil}
	\end{center}
\end{table}

As shown in Table~\ref{tab:gerbil}, NCEL achieves the best performance in most cases with an average gain of 2\% on Micro F1 and 3\% Macro F1. The baseline methods also achieve competitive results on some datasets but fail to adapt to the others. For example, AIDA and xLisa perform quite well on ACE2004 but poorly on other datasets, or WAT, PBoH, and WNED have a favorable performance on CoNLL but lower values on ACE2004 and AQUAINT. Our proposed method performs consistently well on all datasets that demonstrates the good generalization ability.

\subsection{Results on TAC2010 and WW}
\label{sec:tac_ww}
\begin{figure}[htb] 
  \begin{minipage}[b]{0.5\textwidth} 
  \small
    \centering 
    \tabcaption{Precision on WW} 
    \label{table:ww} 
    \begin{tabular}{|c|c|c||c|c|} \hline 
       AIDA & Ganea & WNED & NCEL-local & NCEL\\ \hline 
       0.63 & 0.78 & 0.84 & 0.81 & \textbf{0.86} \\ \hline 
      
    \end{tabular}
    
    \vspace{0.3cm}
    
    \begin{tabular}{|c|c|c|c|} \hline 
        & Prec & Micro F1 & Macro F1 \\ \hline 
      Chisholm & 0.81 & - & - \\ 
      He & 0.81 & - & - \\
      NTEE & 0.88 & - & - \\ \hline
      NCEL-local & 0.89 & 0.89 & 0.88 \\ \hline\hline
      AIDA & - & 0.55 & 0.51 \\
      Babelfy & - & 0.63 & 0.62 \\
      WAT & - & 0.75 & 0.73 \\
      Globerson & - & 0.84 & - \\
      Boosting & - & 0.86 & 0.85 \\ \hline
      NCEL & \textbf{0.91} & \textbf{0.91} & \textbf{0.92} \\ \hline 
    \end{tabular}
    \tabcaption{Results on TAC2010}
    \label{table:tac}
    
    \end{minipage}%
  \begin{minipage}[b]{0.5\textwidth} 
    \centering
    \includegraphics[width=0.9\textwidth]{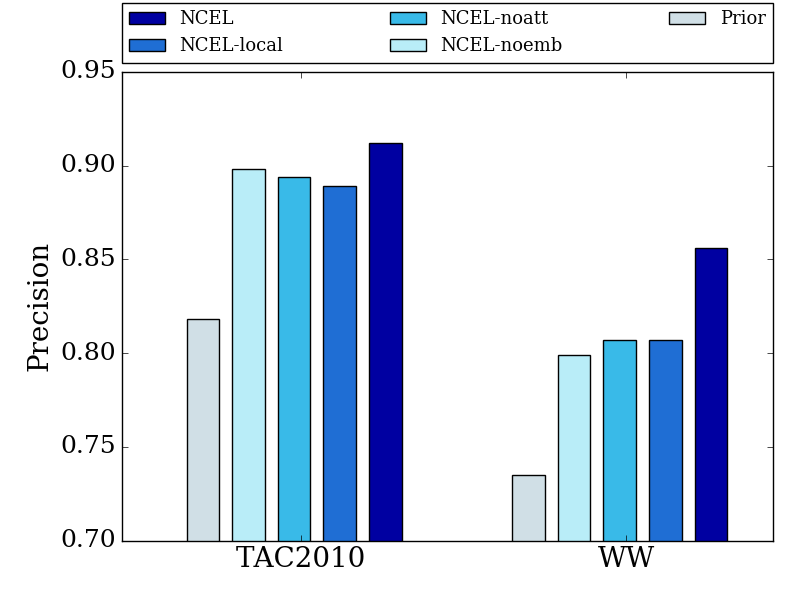}
    \caption{Impacts of NCEL modules} 
    \label{fig:module} 
  \end{minipage}
\end{figure}

In this section, we investigate the effectiveness of NCEL in the ``easy" and ``hard" datasets, respectively. Particularly, TAC2010, which has two mentions per document on average (Section~\ref{sec:data}) and high prior probabilities of correct candidates (Figure~\ref{fig:module}), is regarded as the ``easy" case for EL, and WW is the ``hard" case since it has the most mentions with balanced prior probabilities~\cite{Barbosa2017RobustNE}. Besides, we further compare the impact of key modules by removing the following part from NCEL: global features (NCEL-local), attention (NCEL-noatt), embedding features (NCEL-noemb), and the impact of the prior probability (prior).

The results are shown in Table~\ref{table:ww} and Table~\ref{table:tac}. We can see the average linking precision (Micro) of WW is lower than that of TAC2010, and NCEL outperforms all baseline methods in both easy and hard cases. In the ``easy" case, local models have similar performance with global models since only little global information is available (2 mentions per document). Besides, NN-based models, NTEE and NCEL-local, perform significantly better than others including most global models, demonstrating that the effectiveness of neural models deals with the first limitation in the introduction.

\para{Impact of NCEL Modules}

As shown in Figure~\ref{fig:module}, the prior probability performs quite well in TAC2010 but poorly in WW. Compared with NCEL-local, the global module in NCEL brings more improvements in the ``hard" case than that for ``easy" dataset, because local features are discriminative enough in most cases of TAC2010, and global information becomes quite helpful when local features cannot handle. That is, our propose collective model is robust and shows a good generalization ability to difficult EL. The improvements by each main module are relatively small in TAC2010, while the modules of attention and embedding features show non-negligible impacts in WW (even worse than local model), mainly because WW contains much noise, and these two modules are effective in improving the robustness to noise and the ability of generalization by selecting informative words and providing more accurate semantics, respectively.

\subsection{Qualitative Analysis}
\label{sec:qa}

\begin{table}[htp]
\small
\begin{center}
\begin{tabular}{|c|c|c|c|}
\hline
\multicolumn{4}{|p{15cm}|}{Hussain, considered surplus to {\color{red}{England}}’s one-day requirements, struck 158, his first championship century of the season, as {\color{red}{Essex}} reached 372 and took a first innings lead of 82.} \\ \hline
\multicolumn{2}{|C{7.5cm}}{NCEL} & \multicolumn{2}{|C{7.5cm}|}{NCEL-local} \\ \hline
 England & 0.23 & England & 0.42 \\
 England cricket team & 0.72 & England cricket team & 0.20 \\ \hline
 Essex County Cricket Club & 0.99 & Essex County Cricket Club & 0.97 \\ \hline
\end{tabular}
\caption{Qualitative Analysis of the Example \textit{England}.}
\label{tab:qa}
	\end{center}
\end{table}

The results of example in Figure~\ref{fig:example} are shown in Table~\ref{tab:qa}, which is from CoNLL testa dataset. For mention \textit{Essex}, although both NCEL and NCEL-local correctly identify entity \textit{Essex County Cricket Club}, NCEL outputs higher probability due to the enhancement of neighbor mentions. Moreover, for mention \textit{England}, NCEL-local cannot find enough disambiguation clues from its context words, such as \textit{surplus} and \textit{requirements}, and thus assigns a higher probability of 0.42 to the country \textit{England} according to the prior probability. Collectively, NCEL correctly identifies England cricket team with a probability of 0.72 as compared with 0.20 in NCEL-local with the help of its neighbor mention \textit{Essex}.

\section{Conclusion}
\label{sec:con}
In this paper, we propose a neural model for collective entity linking that is end-to-end trainable. It applies GCN on subgraphs instead of the entire entity graph to efficiently learn features from both local and global information. We design an attention mechanism that endows NCEL robust to noisy data. Trained on collected Wikipedia hyperlinks, NCEL outperforms the state-of-the-art collective methods across five different datasets. Besides, further analysis of the impacts of main modules as well as qualitative results demonstrates its effectiveness.

In the future, we will extend our method into cross-lingual settings to help link entities in low-resourced languages by exploiting rich knowledge from high-resourced languages, and deal with NIL entities to facilitate specific applications.

\section{Acknowledgments}
The work is supported by National Key Research and Development Program of China (2017YFB1002101), NSFC key project (U1736204, 61661146007), and THUNUS NExT Co-Lab.
\bibliographystyle{acl}
\bibliography{coling2018}

\end{document}